\pdfoutput=1
\PassOptionsToPackage{authoryear,round,sort}{natbib}
\documentclass[10pt, logo, copyright]{nv}

\usepackage{booktabs}
\usepackage{caption}
\usepackage{xcolor}
\usepackage{multirow}
\usepackage{makecell}
\usepackage{array}
\usepackage{tabularx}
\usepackage{url}
\usepackage{graphicx}
\usepackage{xspace}
\usepackage{natbib}
\setcitestyle{authoryear,round,semicolon,aysep={,},yysep={;},notesep={, }}
\usepackage{amsmath,amsfonts,bm}
\usepackage{amssymb}
\usepackage{amsthm}
\usepackage{algorithm}
\usepackage{algpseudocode}
\usepackage{pifont} 
\usepackage{tikz}
\usetikzlibrary{tikzmark,calc,positioning,arrows.meta}
\usepackage{wrapfig}
\usepackage{subcaption}

\definecolor{cvprblue}{rgb}{0.21,0.49,0.74}
\definecolor{iccvblue}{rgb}{0.21,0.49,0.74}
\definecolor{mitblue}{rgb}{0.88,0.95,0.96}
\definecolor{gold}{rgb}{0.75,0.6,0.12}
\colorlet{shadecolor}{gray!40}
\definecolor{mydarkred}{rgb}{0.8,0.02,0.02}

\newcolumntype{g}{>{\columncolor{mitblue}}c}
\newcolumntype{f}{>{\columncolor{mitblue}}l}
\newcolumntype{h}{>{\columncolor{mitblue}}r}
\newcolumntype{i}{>{\columncolor{gray}}c}

\newcommand{\cmark}{\ding{51}}%
\newcommand{\xmark}{\ding{55}}%

\newcommand{\framework}{Lightning OPD 2.0\xspace}
\newcommand{\offlinevariant}{\ensuremath{^{\lozenge}}}

\theoremstyle{definition}  

\usepackage[most]{tcolorbox}

\usepackage[
    colorlinks=true,
    citecolor=nvidiagreen,
    linkcolor=nvidiagreen
]{hyperref}

\title{Lightning OPD 2.0: Mitigating Style Bias in Cross-Teacher On-Policy Distillation for Large Reasoning Models}

\author{
Yecheng Wu, Song Han, Han Cai \\~\\
NVIDIA \\
\url{https://github.com/jet-ai-projects/Lightning-OPD}
}

\correspondingauthor{Han Cai (\texttt{hcai@nvidia.com}).}

\begin{document}
\begin{abstract}
On-policy distillation (OPD) provides dense token-level supervision from a teacher, but its effectiveness can depend on \emph{teacher consistency}, meaning that the model providing OPD supervision should also have generated the demonstrations used to train the supervised fine-tuning (SFT) reference. However, this condition is frequently violated in practice when SFT data have mixed or unknown provenance or when different models are preferred for SFT data generation and subsequent distillation. In such cross-teacher settings, even a stronger OPD teacher can yield little improvement over the SFT reference. We find that raw teacher--reference disagreement contains potentially useful context-specific teacher evidence as well as a recurring component associated with differences in wording, formatting, and reasoning cadence. We introduce \textbf{\framework} with cross-fitted style residualization, which uses rollout-level cross-fitting to estimate this recurring component as an operational proxy for style-token bias and subtracts it before constructing the token-level OPD update. Across mathematical reasoning and code generation benchmarks, \framework consistently outperforms Lightning OPD in cross-teacher settings. Starting from Klear-Reasoner-8B-SFT, \framework reaches $82.4\%$ on AIME 2024 and $63.0\%$ on LiveCodeBench v5. Together, these results establish \framework as a practical approach to cross-teacher OPD, relaxing teacher consistency as a prerequisite and allowing the SFT data generator and distillation teacher to be selected independently. Code will be released soon.
\end{abstract}

\maketitle

\begin{figure*}[ht]
    \centering
    \includegraphics[width=\linewidth]{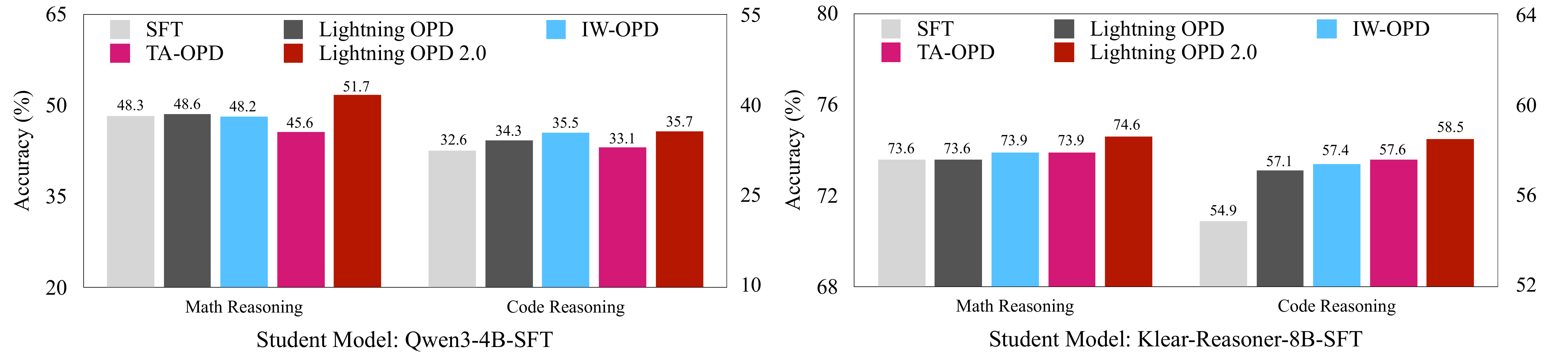}
    \caption{Average Pass@1 (\%) on mathematical reasoning and code generation in two cross-teacher settings with Qwen3-30B-A3B-Thinking-2507 as the OPD teacher. \framework achieves the strongest average performance across both task families and both SFT references, consistently improving over Lightning OPD.}
    \label{fig:teaser}
\end{figure*}

\section{Introduction}
\label{sec:intro}

On-policy distillation (OPD) has become an effective post-training alternative to reinforcement learning for large language models (LLMs) \citep{agarwal2024policy,lu2025onpolicydistillation,yang2026learning,Nemotron_Cascade_2}. Rather than learning from a sparse sequence-level reward, OPD utilizes a high-capability teacher to score student-generated trajectories and converts teacher--student disagreement into dense token-level supervision. This dense on-policy supervision makes OPD a stable and cost-effective alternative to reinforcement learning from verifiable rewards (RLVR), while maintaining strong performance across reasoning tasks \citep{lu2025onpolicydistillation,yang2025qwen3,yang2026learning,Nemotron_Cascade_2,zeng2026glm5,xiao2026mimo}.

Standard OPD nevertheless requires a live teacher throughout training. \citet{wu2026lightning} introduce Lightning OPD, which removes this systems bottleneck by precomputing both the rollouts from the supervised fine-tuning (SFT) reference policy and their corresponding teacher log-probabilities before training, then reusing the resulting cache throughout optimization. A central finding of that work is that effective OPD requires \emph{teacher consistency}. This condition means that the teacher used for OPD is the same model that generated the demonstrations used to train the SFT reference policy. Teacher consistency aligns the token-level distillation signal with the behavior already learned by the reference policy, which allows fully offline Lightning OPD to closely track standard online OPD. This consideration is not unique to the offline formulation. Standard OPD also begins from an SFT policy shaped by its demonstration teacher. Selecting a different OPD teacher can therefore introduce the cross-teacher mismatch even though rollouts and teacher scores are collected online.

However, teacher consistency is often difficult to satisfy in practice. Available SFT data may have mixed or undocumented provenance and may have been generated by several different models, leaving no single demonstration teacher that can be reused consistently during OPD. Even when the original data generator is known, practitioners may want to reuse an existing SFT dataset rather than incur the substantial cost of regenerating demonstrations with every candidate teacher and repeating SFT. More fundamentally, teacher consistency couples the choice of the SFT data generator with the choice of the OPD teacher, even though the best models for these two roles need not be the same. The intended OPD teacher may not be the most suitable model for generating SFT demonstrations. Practitioners may instead use a stronger generator to produce higher-quality demonstrations and obtain a stronger SFT reference policy, while selecting a different teacher for later distillation to maximize the overall post-training result. Enforcing consistency in these settings would require rebuilding the SFT pipeline or compromising one stage to accommodate the other. The practical goal is therefore to preserve the strongest available SFT reference while retaining the freedom to choose the most suitable teacher for subsequent OPD.

To this end, our goal is to remove teacher consistency as a practical prerequisite for effective OPD and decouple the choice of the SFT data generator from that of the OPD teacher while still achieving reliable post-training performance across different teacher combinations. However, naively replacing the consistent teacher with a different OPD teacher substantially degrades performance, even when the new teacher is stronger.

To understand this degradation, we consider the role of each token in the response. Some tokens carry problem-specific reasoning, whereas others primarily express the response's style. On reasoning-related tokens, a low probability from the OPD teacher may identify an incorrect intermediate conclusion, operation, or reasoning direction and thereby provide useful corrective supervision. On style tokens, however, the same low probability may instead reflect a preference for different wording, transitions, formatting, derivation length, or reasoning cadence, even when the reference trajectory remains valid. Lightning OPD cannot distinguish these cases and converts both into the same token-level update. Because style choices appear throughout a response and similar preferences recur across rollouts, their accumulated penalties may systematically interfere with the more context-specific reasoning signal. This difference also suggests an observable statistical proxy. Disagreement associated with style tokens is more likely to recur across unrelated rollouts through similar lexical choices, response positions, and levels of reference-policy surprisal, whereas reasoning-related disagreement depends more strongly on the current problem and reasoning state. We therefore treat the component that is predictable across cached rollouts as a proxy for style-token bias and the remaining residual as a candidate carrier of reasoning-related teacher evidence. These terms describe the roles that motivate our method rather than a ground-truth semantic partition, and our procedure does not assume that every predictable effect is style or that every residual reflects correct reasoning.

Based on this observation, we introduce \textbf{\framework} with cross-fitted style residualization. We first compute the token-level log-probability difference between the OPD teacher and the SFT reference. We then split the cached rollouts into folds and use the other folds to construct two lookup tables, one indexed by token identity and the other by normalized response position and reference-policy surprisal. Averaging the two lookup values gives each held-out token an estimate of its recurring style bias without using its own rollout. We subtract this estimate from the raw difference and use the residual in place of the original disagreement in the Lightning OPD objective.

Using Qwen3-30B-A3B-Thinking-2507 as the OPD teacher, we evaluate \framework on mathematical reasoning and code generation benchmarks with Qwen3-4B-SFT and Klear-Reasoner-8B-SFT as two distinct SFT references. \framework consistently outperforms Lightning OPD in both settings. It improves average mathematical reasoning performance by $3.1$ and $1.0$ points, respectively, and average code generation performance by $1.4$ points in both settings. Starting from Klear-Reasoner-8B-SFT, \framework reaches $82.4\%$ on AIME 2024 and $63.0\%$ on LiveCodeBench v5.

Our contributions are threefold.
\begin{itemize}
    \item We formulate the \emph{cross-teacher} OPD setting, in which the SFT data generator and the OPD teacher are selected independently, and show that distillation effects can substantially degrade under this setting.

    \item We identify a recurring component of teacher--reference disagreement that is predictable across  rollouts and associated with style-token bias, explaining why raw cross-teacher supervision can be misleading.

    \item We introduce \framework, which estimates this recurring component with cross-fitted token and context lookup tables and subtracts it before applying the Lightning OPD update. Across two cross-teacher settings, \framework consistently improves Lightning OPD on mathematical reasoning and code generation benchmarks and achieves state-of-the-art performance.
\end{itemize}

\section{Related Work}
\label{sec:related}

\paragraph{LLM Post-Training.}
Post-training is central to capable LLMs, commonly combining supervised fine-tuning on high-quality instruction and reasoning demonstrations \citep{ouyang2022training,guha2025openthoughts,yang2024qwen25math,guo2025deepseek} with preference- and reward-based optimization \citep{schulman2017proximal,rafailov2023direct,ahmadian2024back,li2023remax,dong2023raft}. For reasoning, recent methods develop outcome-based policy optimization \citep{shao2024deepseekmath,hu2025reinforcepp,liu2025drgrpo,yu2025dapo,zheng2025gspo,minimax2025cispo} and improve process-level credit assignment or value learning \citep{cui2025prime,kazemnejad2024vineppo,yuan2025vcppo,yue2025vapo}. These advances support increasingly capable mathematical and general reasoning systems \citep{guo2025deepseek,liu2024deepseekv3,team2025kimik15,hu2025openreasoner,yang2024qwen25math,yang2025qwen3,Nemotron_Cascade,Nemotron_Cascade_2,nvidia_nemotron_3_2025,xiao2026mimo,team2026kimi,singh2025openai,zeng2026glm5}. Complementary analyses study what online optimization adds beyond SFT and why it can better preserve existing capabilities \citep{yue2025does,shenfeld2025razor}, while recent methods co-design or interleave SFT and RL \citep{yan2025luffy,ma2026relift,zhang2025chord,chen2025bridge,liu2025uft,huang2025blend}. Our work instead studies the coupling between the model that generates the SFT demonstrations and the teacher that provides subsequent OPD supervision, seeking to remove teacher consistency as a practical prerequisite for effective OPD.

\paragraph{On-Policy Distillation.}
Knowledge distillation transfers capabilities from a teacher to a student by matching their output distributions \citep{hinton2015distilling}. Sequence-level and language-model distillation extend this idea to autoregressive generation \citep{kim2016sequence,gu2024minillm,rang2025revealing,xu2025speculative}, whereas on-policy distillation evaluates the teacher on student-generated trajectories and provides dense feedback on states visited by the student \citep{agarwal2024policy,lu2025onpolicydistillation,song2026survey}. Recent work extends OPD through reward extrapolation and flexible reference policies \citep{yang2026learning}, entropy-aware divergence and local-support matching \citep{wang2026entropy,fu2026revisiting,ke2026respecting}, reinforcement-aware or verifier-guided objectives \citep{xu2026reinforcement,xu2026sign}, controllable reasoning \citep{liang2026orbit}, and black-box, privileged, or self-distillation settings \citep{ye2025blackbox,zhao2026self,shenfeld2026self,hubotter2026reinforcement,penaloza2026privileged,tan2026selfsupervised,yang2026ogls}. OPD has also become a component of large-scale reasoning post-training pipelines \citep{yang2025qwen3,xiao2026mimo,Nemotron_Cascade_2}. Complementary studies show that stronger teachers do not always yield better students and trace failures to thinking-pattern incompatibility, unreliable token-level guidance, and teacher-dependent gradient quality \citep{chen2026rethinking,fu2026revisiting,armandpour2026unmasking,xu2026sign,yang2026ogls}. Offline OPD variants avoid maintaining a live teacher during optimization by precomputing rollouts and their corresponding teacher log-probabilities before training \citep{rang2025revealing,wu2026lightning}. Building upon Lightning OPD, we study how to maintain effective distillation in cross-teacher settings, where the SFT data generator and the OPD teacher differ. \framework uses rollout-level cross-fitting to estimate the recurring component of teacher--reference disagreement as an operational proxy for style-token bias and subtracts it before constructing the token-level OPD update.

\section{Methodology}
\label{sec:method}

\subsection{Cross-Teacher OPD}
\label{subsec:problem}

Let $p$ be the training-prompt distribution and $\{q_i\}_{i=1}^{N}$ the prompts stored in the replay. When the supervised fine-tuning (SFT) demonstrations have a single identifiable generator, let $\pi_G$ denote this SFT data generator. Let $\pi_R$ be the resulting SFT reference policy, $\pi_T$ the model selected to provide OPD supervision, and $\pi_\theta$ a trainable policy initialized from $\pi_R$. Teacher consistency requires $\pi_T=\pi_G$~\citep{wu2026lightning}. We study \emph{cross-teacher OPD}, where $\pi_T$ is selected independently of $\pi_G$ or no single $\pi_G$ can be identified. Our method addresses this general setting through the frozen replay of Lightning OPD.

Following Lightning OPD, we sample and freeze one response $x_i=(y_{i1},\ldots,y_{iL_i})\sim\pi_R(\cdot\mid q_i)$ for every prompt. Let $h_{it}=(q_i,y_{i,<t})$ be the prefix at token $t$. The OPD teacher and the SFT reference score the same realized token, producing the cached chosen-token log-probabilities
\begin{equation}
    \ell^T_{it}=\log\pi_T(y_{it}\mid h_{it}),
    \qquad
    \ell^R_{it}=\log\pi_R(y_{it}\mid h_{it}).
\end{equation}
We assume that these scores are aligned to the same response tokens. Our experiments use models with compatible Qwen-family tokenization.

Using this cache, unmodified Lightning OPD retains the original per-token advantage
\begin{equation}
    A^{\mathrm{base}}_{it}(\theta)
    =\ell^T_{it}-\log\pi_\theta(y_{it}\mid h_{it}),
    \label{eq:base_advantage}
\end{equation}
and optimizes
\begin{equation}
    J_{\mathrm{base}}(\theta)
    =\mathbb E_{q_i\sim p,\,x_i\sim\pi_R}
    \!\left[\sum_{t=1}^{L_i}A^{\mathrm{base}}_{it}(\theta)\right].
    \label{eq:base_objective}
\end{equation}
Following standard OPD practice~\citep{agarwal2024policy,lu2025onpolicydistillation,wu2026lightning}, the advantage is treated as a fixed scalar when computing parameter updates. Throughout this section, $\nabla J$ denotes the resulting advantage-weighted policy gradient rather than ordinary differentiation through the advantage. We use the same policy surrogate as Lightning OPD.

To isolate the part of the signal affected by the teacher choice, we define the teacher--reference disagreement
\begin{equation}
    d_{it}=\ell^T_{it}-\ell^R_{it}.
    \label{eq:raw_disagreement}
\end{equation}
At initialization, $\pi_\theta=\pi_R$, so $A^{\mathrm{base}}_{it}=d_{it}$. In cross-teacher OPD, $d_{it}$ can contain both context-specific teacher evidence and recurring teacher--reference differences induced by the mismatch with the SFT data generator. The unmodified update cannot distinguish the two. The next subsection develops an operational proxy for this recurring component.

\subsection{Predictability across Rollouts as a Style Proxy}
\label{subsec:style_proxy}

\begin{figure*}[t]
    \centering
    \includegraphics[width=\linewidth]{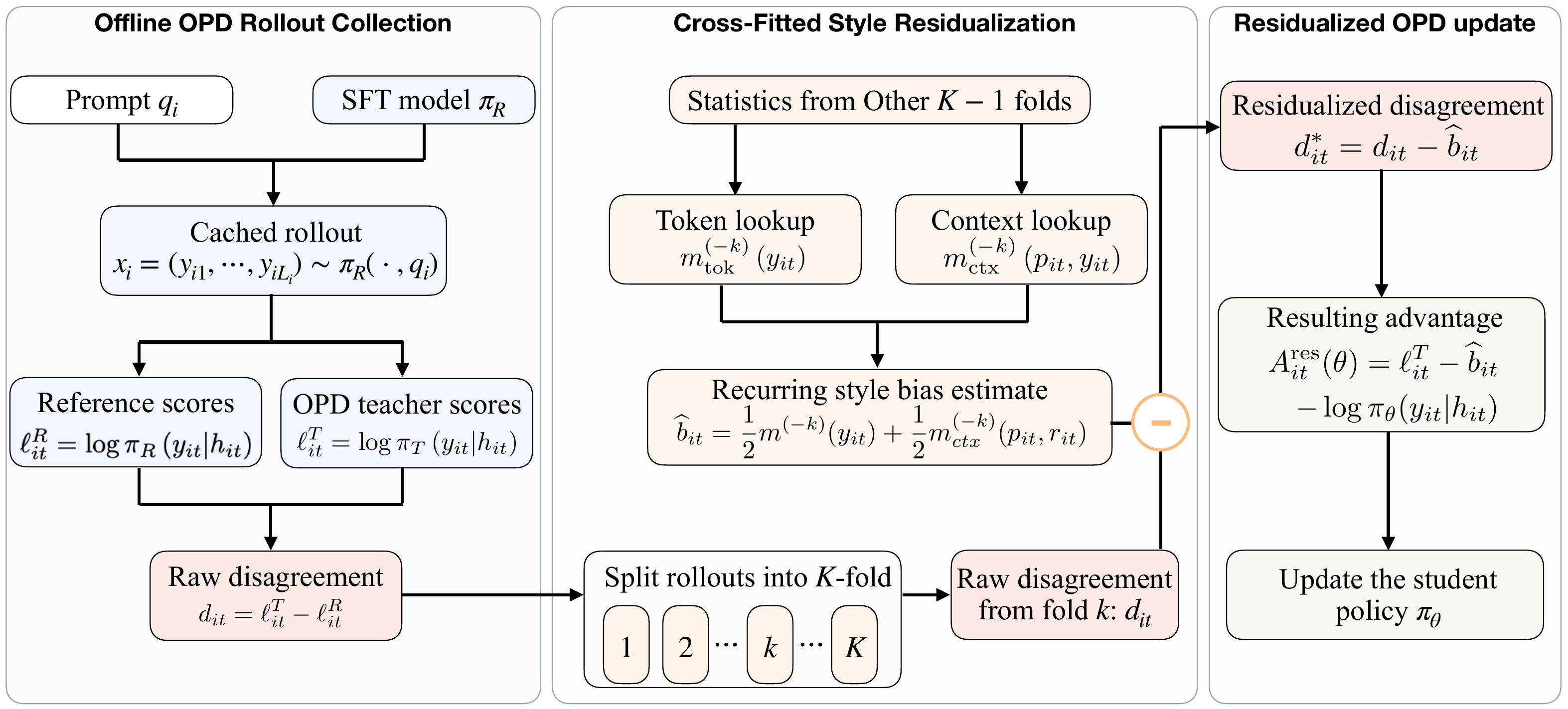}
    \caption{
    \looseness=-1
    Overview of \framework. The SFT reference generates the frozen rollouts, which are scored by both the reference and the independently selected OPD teacher to obtain the raw teacher--reference disagreement. For each held-out fold, response-balanced token and context lookup tables fitted on the other folds estimate the recurring disagreement as an operational proxy for style-token bias. Subtracting this estimate yields a residualized signal for the reference-anchored Lightning OPD update.}
    \label{fig:overview}
\end{figure*}

The raw disagreement $d_{it}$ can contain both context-specific teacher evidence and systematic differences in how the selected teacher and the SFT reference express a response. We make the operational assumption that disagreement associated with recurring style differences is more predictable across cached rollouts from simple lexical and coarse contextual coordinates, whereas potentially useful, reasoning-related teacher evidence is more dependent on the current context. We therefore write
\begin{equation}
    d_{it}=b(z_{it})+v_{it},
    \label{eq:decomposition}
\end{equation}
where $z_{it}$ collects the token and context coordinates defined below, $b(z_{it})$ is the component that recurs at those coordinates across rollouts, and $v_{it}$ is the remaining variation. This is an operational decomposition rather than a semantic labeling of individual tokens. We use $b(z_{it})$ as a proxy for recurring style-token bias and approximate it by equally combining token- and context-based lookup predictions. We estimate these predictions by rollout-level cross-fitting and replace $d_{it}$ with its estimated residual in the training objective.

\paragraph{Observable coordinates for recurring disagreement.}
Estimating a component that recurs across rollouts requires coordinates that are shared often enough to support reliable averaging but expressive enough to distinguish different uses of a token. Token identity captures lexical preference but treats every occurrence alike. At the other extreme, the full prefix identifies the local context precisely but rarely repeats across rollouts and may absorb problem-specific reasoning. We therefore complement token identity with a coarse context coordinate based on normalized response position and reference-policy surprisal. Normalized response position identifies the stage of the response. Reference-policy surprisal provides a low-dimensional summary of how typical the realized token is under the SFT reference, rather than a direct indicator of style. Together, these coordinates describe broad usage conditions without conditioning on the full prompt or prefix.

Formally, let the reference-policy surprisal be $\xi_{it}=-\ell^R_{it}$, the standard measure of how atypical a realized token is under a language model~\citep{wilcox2023testing}. A small value means that the token is natural under the SFT reference, while a large value means that the reference itself finds the token unexpected. We divide the response into $B_{\mathrm{pos}}$ normalized-position bins and the surprisal range into $B_{\mathrm{ref}}$ bins up to $\xi_{\max}$. Coarse surprisal bins distinguish disagreement on reference-typical and reference-atypical tokens while preserving sufficient support across rollouts for stable estimation. Their zero-indexed identifiers are
\begin{equation}
    p_{it}=\left\lfloor B_{\mathrm{pos}}\frac{t-1}{L_i}\right\rfloor,
    \qquad
    r_{it}=\min\!\left\{B_{\mathrm{ref}}-1,
    \left\lfloor B_{\mathrm{ref}}\frac{\xi_{it}}{\xi_{\max}}\right\rfloor\right\}.
    \label{eq:nuisance_coordinates}
\end{equation}
We set $z^{\mathrm{tok}}_{it}=y_{it}$, $z^{\mathrm{ctx}}_{it}=(p_{it},r_{it})$, and $z_{it}=(z^{\mathrm{tok}}_{it},z^{\mathrm{ctx}}_{it})$. Binning pools comparable occurrences across responses of different lengths. We construct the token and context lookup tables separately rather than using their full cross product, which avoids fragmenting the cache into excessively sparse groups.

\subsection{Cross-Fitted Estimation of Recurring Disagreement}
\label{subsec:style_estimation}

Cross-fitting is a standard sample-splitting technique for constructing held-out predictions and preventing self-fitting~\citep{chernozhukov2018double,guo2021machine}. We use it only as an estimation device. We partition the cached rollouts into $K$ deterministic folds, keeping all rollouts originating from the same prompt in one fold. For a rollout in fold $k$, we estimate all statistics using only the other $K-1$ folds. Neither that rollout nor another rollout from the same prompt can therefore contribute to its lookup tables. This exclusion prevents self-fitting but does not by itself identify the predicted component as style.

To prevent longer responses from dominating the estimates, we assign each response unit total weight, distributed uniformly over its active tokens, when fitting the lookup tables. The training loss uses the same reduction by averaging active-token contributions within each response and then averaging across responses. This response-balanced reduction is shared by all methods and is not a component of style residualization.

Using the non-held-out folds, we build one lookup table for the mean disagreement of each token identity and another for each context coordinate. Rare groups are smoothed toward the global mean of the non-held-out folds, while unseen groups use that global mean directly. This prevents a rare token or context from receiving an unstable correction. Let $m_{\mathrm{tok}}^{(-k)}$ and $m_{\mathrm{ctx}}^{(-k)}$ denote the resulting smoothed lookup tables. For token $(i,t)$ in held-out fold $k$, we use the equal-weight prediction
\begin{equation}
    \widehat b_{it}
    =\frac{1}{2}m_{\mathrm{tok}}^{(-k)}(z^{\mathrm{tok}}_{it})
    +\frac{1}{2}m_{\mathrm{ctx}}^{(-k)}(z^{\mathrm{ctx}}_{it}).
    \label{eq:nuisance_predictor}
\end{equation}
The two lookup predictions receive equal coefficients, and averaging keeps their combination on the scale of a single disagreement estimate. Equation~\eqref{eq:nuisance_predictor} is our concrete approximation to $b(z_{it})$ in Equation~\eqref{eq:decomposition}. At the conceptual level, both tables average the raw disagreement $d_{it}$. Because both lookup tables exclude the current rollout, $\widehat b_{it}$ captures recurring disagreement that is predictable from other cached rollouts rather than variation fitted specifically to the current response.

\subsection{Residualized Lightning OPD}
\label{subsec:residualized_opd}

The base advantage separates into the cross-teacher disagreement and a reference-anchoring term
\begin{equation}
    A^{\mathrm{base}}_{it}(\theta)
    =d_{it}+\ell^R_{it}-\log\pi_\theta(y_{it}\mid h_{it}).
    \label{eq:base_decomposition}
\end{equation}
Because $\widehat b_{it}$ estimates recurring teacher--reference disagreement, we residualize only $d_{it}$ and leave the reference-anchoring term unchanged. We set
\begin{equation}
    d^*_{it}=d_{it}-\widehat b_{it}.
    \label{eq:corrected_score}
\end{equation}
The resulting advantage is
\begin{equation}
    \begin{aligned}
        A^{\mathrm{res}}_{it}(\theta)
        &=d^*_{it}+\ell^R_{it}-\log\pi_\theta(y_{it}\mid h_{it}) \\
        &=\ell^T_{it}-\widehat b_{it}-\log\pi_\theta(y_{it}\mid h_{it}).
    \end{aligned}
    \label{eq:cfsr_advantage}
\end{equation}
At initialization, $\pi_\theta=\pi_R$ and $A^{\mathrm{res}}_{it}=d^*_{it}$. We retain the Lightning OPD objective form while replacing the raw cross-teacher disagreement with its residualized counterpart
\begin{equation}
    J_{\mathrm{res}}(\theta)
    =\mathbb E_{q_i\sim p,\,x_i\sim\pi_R}
    \!\left[\sum_{t=1}^{L_i}A^{\mathrm{res}}_{it}(\theta)\right].
    \label{eq:cfsr_objective}
\end{equation}
Equivalently, at the conceptual objective level, we define the fixed effective chosen-token teacher score $\widetilde\ell^T_{it}=\ell^T_{it}-\widehat b_{it}$ and use it in place of $\ell^T_{it}$. This score and $\widehat b_{it}$ are computed before training and remain fixed throughout optimization.

\begin{algorithm}[t]
\caption{Lightning OPD 2.0 with Cross-Fitted Style Residualization}
\label{alg:cfsr}
\begin{algorithmic}[1]
\Require SFT reference $\pi_R$, frozen replay $\mathcal D$ with scores $\ell^R$ and $\ell^T$, and number of folds $K$
\State Compute $d_{it}$ and the token and context coordinates defined in Section~\ref{subsec:style_proxy}
\State Partition rollouts into $K$ folds, grouping those from the same prompt
\For{each held-out fold $k$}
    \State Fit response-balanced, smoothed tables $m_{\mathrm{tok}}^{(-k)}$ and $m_{\mathrm{ctx}}^{(-k)}$ on the other folds
    \State Correct fold $k$ using Equation~\eqref{eq:nuisance_predictor}, setting $\widetilde\ell^T_{it}=\ell^T_{it}-\widehat b_{it}$
\EndFor
\State Initialize $\pi_\theta\leftarrow\pi_R$
\For{each training step}
    \State Compute $A^{\mathrm{res}}_{it}$ on a minibatch from the corrected replay using Equation~\eqref{eq:cfsr_advantage}
    \State Update $\theta$ with the Lightning OPD policy surrogate
\EndFor
\State \Return $\pi_\theta$
\end{algorithmic}
\end{algorithm}

\section{Experiments}
\label{sec:exp}

Our experiments are organized into four parts. We first describe the two cross-teacher settings, training and evaluation protocols, and matched baselines. We then report the main results on mathematical reasoning and code generation under both settings. Next, we conduct a mechanism analysis that compares the cross-teacher training signal with its teacher-consistent counterpart and shows how \framework removes the dominant style-token bias introduced by teacher inconsistency. Finally, we ablate the core algorithmic design points by removing the token-identity lookup, the context lookup, and prompt-level cross-fitting from the full method.

\subsection{Experimental Setup}
\label{subsec:setup}

\paragraph{Cross-teacher settings.}
We study two complementary settings. The Qwen3-4B setting starts from the Qwen3-4B SFT reference used by Lightning OPD, whose SFT demonstrations were generated by Qwen3-8B, and selects Qwen3-30B-A3B-Thinking-2507 as the OPD teacher. Its known provenance makes the source of cross-teacher style mismatch directly identifiable. The second setting uses Klear-Reasoner-8B-SFT~\citep{kwai2026klear}, which is trained from Qwen3-8B-Base using long-chain-of-thought SFT data distilled from DeepSeek-R1-0528~\citep{deepseek2025r10528}. It tests whether the same correction applies to a stronger SFT reference trained from a different initialization and data generator. Both settings use Qwen3-30B-A3B-Thinking-2507~\citep{yang2025qwen3} as the selected cross-teacher OPD teacher.

\paragraph{Training Settings.}
We train on two domains. Mathematical reasoning uses DAPO-Math-17k~\citep{yu2025dapo}, which provides 17K competition-level math problems spanning a wide range of difficulty. Code generation uses KlearReasoner-CodeSub-15K~\citep{kwai2026klear}, which provides 15K carefully cleaned and filtered code problems. For each OPD prompt, we sample a single response from the SFT reference and precompute the corresponding selected-teacher log-probabilities once before training, following the pipeline of Lightning OPD~\citep{wu2026lightning}. We use five prompt-level folds for cross-fitting. The training framework is built upon slime~\citep{slime2025}, and we train each model for 150 steps using the Lightning OPD policy surrogate with a PPO clipping range of $0.2$.

\paragraph{Evaluation.}
For mathematical reasoning, we evaluate on AIME 2024~\citep{aimo2024aime}, AIME 2025~\citep{opencompass2025aime}, and HMMT February 2025~\citep{balunovic2025matharena}. For code generation, we evaluate on LiveCodeBench v5 and v6~\citep{jain2024livecodebench}. We set the temperature to $0.6$ and top-$p$ to $0.95$ throughout. The maximum generation length is 40,960 for all math and code benchmarks. Math uses 64 solutions per problem and code uses 8 solutions per problem, and we report average Pass@1.

\paragraph{Baselines.}
We compare \framework with the SFT reference and three OPD baselines. Lightning OPD~\citep{wu2026lightning} applies the original token-level distillation objective without style correction. IW-OPD~\citep{xie2026position} addresses position bias by weighting tokens according to the accumulated teacher--student discrepancy, while TA-OPD~\citep{wang2026teachability} selects positions whose teacher signal is predicted to be learnable. For a controlled comparison, all OPD methods start from the same SFT reference and use the same rollouts, cached teacher log-probabilities, and training budget. Since IW-OPD and TA-OPD were originally formulated with online rollouts, we adapt their objectives to the frozen Lightning OPD replay.

\subsection{Main Results}
\label{subsec:main_results}

Table~\ref{tab:main} presents evaluation results across five benchmarks under the two cross-teacher settings. Across both settings, \framework consistently improves the SFT reference, whereas Lightning OPD and the two competing OPD baselines yield little or no aggregate improvement once teacher consistency is violated. In the Qwen3-4B setting, \framework improves all five benchmarks, raising the average math and code scores over the SFT reference by $3.4$ and $3.1$ points, respectively. Compared with Lightning OPD, \framework further improves average math and code performance by $3.1$ and $1.4$ points. It also achieves the strongest average results among all methods, outperforming IW-OPD by $3.5$ points on math and $0.2$ points on code, and TA-OPD by $6.1$ and $2.6$ points. In the Klear-Reasoner-8B-SFT setting, \framework similarly improves all five benchmarks, with average gains of $1.0$ points on math and $3.6$ points on code over the SFT reference. Compared with Lightning OPD, it improves average math and code performance by $1.0$ and $1.4$ points, respectively, and remains stronger than IW-OPD and TA-OPD in both domains. Together, these results demonstrate that \framework generalizes across SFT initializations and data-generation pipelines without requiring teacher consistency.

\begin{table*}[t]
    \centering
    \caption{Pass@1 (\%) on math reasoning and code generation under two cross-teacher settings. Methods marked with $\lozenge$ are frozen-replay adaptations. \textbf{Bold} indicates the strongest result within each setting, and blue rows highlight \framework.}
    \label{tab:main}
    \setlength{\tabcolsep}{3pt}
    \begin{tabular}{lccccccc}
    \toprule
    \multirow{2}{*}{\textbf{Method}}
    & \multicolumn{4}{c}{\textbf{Math Reasoning}} & \multicolumn{3}{c}{\textbf{Code Generation}} \\
    \cmidrule(lr){2-5}\cmidrule(lr){6-8}
    & \textbf{AIME 2024} & \textbf{AIME 2025} & \textbf{HMMT Feb. 2025} & \textbf{Avg.}
    & \textbf{LCB v5} & \textbf{LCB v6} & \textbf{Avg.} \\
    \midrule
    \multicolumn{8}{l}{\emph{Student: Qwen3-4B-SFT\quad Teacher: Qwen3-30B-A3B-Thinking-2507}} \\
    \midrule
    SFT reference               & 57.5 & 52.4 & 34.9 & 48.3 & 33.8 & 31.5 & 32.6 \\
    Lightning OPD               & 59.6 & 51.6 & 34.6 & 48.6 & 35.5 & 33.2 & 34.3 \\
    IW-OPD\offlinevariant       & 60.8 & 49.9 & 33.8 & 48.2 & \textbf{37.2} & 33.8 & 35.5 \\
    TA-OPD\offlinevariant       & 56.0 & 50.0 & 30.6 & 45.6 & 35.7 & 30.4 & 33.1 \\
    \addlinespace[1pt]
    \rowcolor{mitblue}
    \framework                  & \textbf{63.5} & \textbf{55.0} & \textbf{36.7} & \textbf{51.7} & 36.9 & \textbf{34.4} & \textbf{35.7} \\
    \midrule
    \multicolumn{8}{l}{\emph{Student: Klear-Reasoner-8B-SFT\quad Teacher: Qwen3-30B-A3B-Thinking-2507}} \\
    \midrule
    Qwen3-8B                    & 76.0 & 67.3 & 44.7 & 62.7 & 57.5 & 48.4 & 53.1 \\
    SFT reference               & 81.3 & 77.5 & 62.2 & 73.6 & 58.5 & 51.3 & 54.9 \\
    Lightning OPD               & 80.6 & 77.2 & 62.9 & 73.6 & 61.6 & 52.6 & 57.1 \\
    IW-OPD\offlinevariant       & 81.9 & 77.6 & 62.3 & 73.9 & 61.0 & 53.8 & 57.4 \\
    TA-OPD\offlinevariant       & \textbf{82.7} & 76.9 & 62.0 & 73.9 & 62.1 & 53.1 & 57.6 \\
    \addlinespace[1pt]
    \rowcolor{mitblue}
    \framework                  & 82.4 & \textbf{77.7} & \textbf{63.8} & \textbf{74.6} & \textbf{63.0} & \textbf{53.9} & \textbf{58.5} \\
    \bottomrule
    \end{tabular}
\end{table*}

\subsection{Mechanism Analysis of Style-Token Bias}
\label{subsec:style_analysis}

We assess whether residualization moves the cross-teacher training signal toward the signal obtained under teacher consistency. Let $\pi_D$ denote the model used for this post-hoc diagnostic and let $\ell^D_{it}$ be its chosen-token log-probability. For each scored response token in the cache, we measure the absolute deviation of the raw and residualized signals from the diagnostic reference signal
\begin{equation}
    o_{it}
    = \left|d_{it} - \left(\ell^D_{it}-\ell^R_{it}\right)\right|,
    \qquad
    o^*_{it}
    = \left|d^*_{it} - \left(\ell^D_{it}-\ell^R_{it}\right)\right|,
    \label{eq:style_deviation}
\end{equation}
where $d_{it}=\ell^T_{it}-\ell^R_{it}$ is the original cross-teacher disagreement and $d^*_{it}$ is its residualized counterpart. In the Qwen3-4B-SFT setting, $\pi_D$ is Qwen3-8B, the provenance teacher $\pi_G$ that generated the SFT demonstrations. The diagnostic reference is therefore the exact teacher-consistent signal in this setting. Exact teacher-consistent scores are unavailable in the Klear-Reasoner-8B-SFT setting, so we use Klear-Reasoner-8B~\citep{kwai2026klear} as a diagnostic proxy. Neither diagnostic model contributes to correction estimation or student training.

\begin{figure*}[t]
    \centering
    \includegraphics[width=\linewidth]{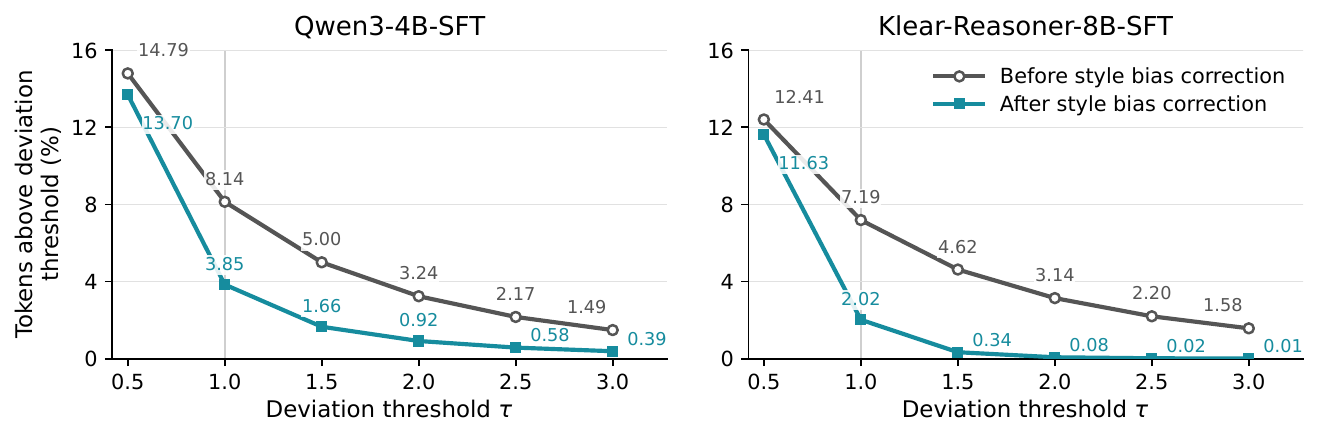}
    \caption{Absolute-deviation threshold analysis. Each curve reports the fraction of scored response tokens whose absolute deviation from the diagnostic reference signal exceeds $\tau$, using $o_{it}$ before style-bias correction and $o^*_{it}$ after correction. The vertical line at $\tau=1$ marks the representative reference point used in the text. Qwen3-8B provides the exact teacher-consistent reference for Qwen3-4B-SFT, while Klear-Reasoner-8B provides a diagnostic proxy for Klear-Reasoner-8B-SFT. Both models are used only for this post-hoc analysis.}
    \label{fig:deviation_threshold_sweep}
\end{figure*}

Figure~\ref{fig:deviation_threshold_sweep} compares the empirical fractions of tokens satisfying $o_{it}>\tau$ and $o^*_{it}>\tau$ as $\tau$ varies from $0.5$ to $3.0$ nats. The after-correction curve lies below the before-correction curve at every evaluated threshold in both settings, showing that the result is not specific to a single cutoff. At $\tau=1$, which we use as a representative reference point, residualization reduces the exceedance rate from $8.14\%$ to $3.85\%$ in the Qwen3-4B-SFT setting and from $7.19\%$ to $2.02\%$ in the Klear-Reasoner-8B-SFT setting. These changes correspond to relative reductions of $52.8\%$ and $71.9\%$, respectively. The relative reduction grows at higher thresholds, showing that residualization reduces the prevalence of large deviations from the diagnostic reference signal.

\begin{figure*}[htbp]
    \centering

    \includegraphics[width=\linewidth]{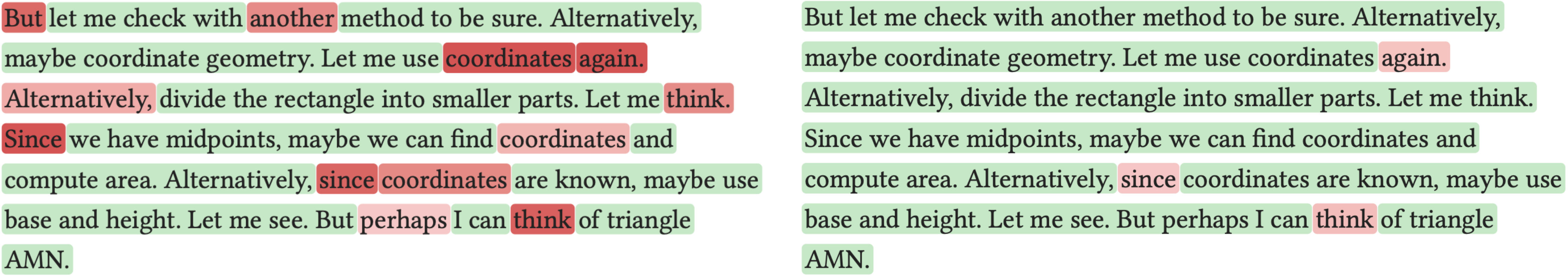}\par
    \vspace{1.5mm}

    \includegraphics[width=\linewidth]{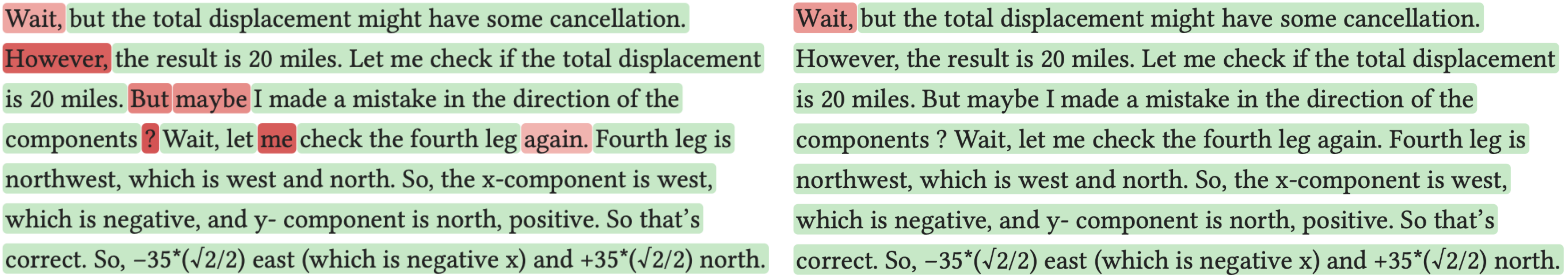}\par
    \caption{Token-level examples of style-bias correction in the Qwen3-4B-SFT setting. Each row shows one cached response before correction on the left and after residualization on the right. Green denotes an absolute deviation of at most one nat from the diagnostic reference signal. Red denotes a deviation above one nat, with darker shades indicating larger values. The one-nat cutoff is used only for visualization.}
    \label{fig:style_token_examples}
\end{figure*}

Figure~\ref{fig:style_token_examples} provides a complementary token-level view using two cached responses from the Qwen3-4B-SFT setting. For visualization, we color tokens with deviations of at most one nat in green and those above one nat in red, with darker shades denoting larger deviations. In these examples, several large raw deviations occur on discourse markers and reasoning transitions. Most fall below the reference threshold after residualization, although some large deviations remain. The examples qualitatively illustrate the threshold-sweep result without treating the visualization threshold as part of the method. Together, the aggregate and token-level evidence supports our operational interpretation of the removed component as recurring style-token bias.

\subsection{Ablation Study}
\label{subsec:ablations}

Table~\ref{tab:component_ablation} ablates the three principal design choices in \framework. The context-only and token-only variants retain one lookup at unit weight, whereas the in-sample variant retains both lookups but removes prompt-level cross-fitting. Lightning OPD serves as the uncorrected baseline, and all variants otherwise share the same cache and training configuration within each setting. The full method yields the strongest overall reported point estimates. The effects of using a single coordinate family vary across settings, while in-sample estimation is generally weaker or tied. These results support combining lexical and coarse contextual estimates and retaining cross-fitting primarily as a safeguard against self-fitting.

\begin{table*}[t]
    \centering
    \caption{Component ablations of \framework on AIME 2024 and HMMT February 2025. All entries report Pass@1 in percent. Checkmarks indicate retained components, and boldface marks the highest point estimate in each column.}
    \label{tab:component_ablation}
    \small
    \setlength{\tabcolsep}{4.5pt}
    \begin{tabular}{lccccccc}
        \toprule
        \multirow{2}{*}{\textbf{Variant}}
        & \multicolumn{3}{c}{\textbf{Components}}
        & \multicolumn{2}{c}{\textbf{Qwen3-4B-SFT}}
        & \multicolumn{2}{c}{\textbf{Klear-Reasoner-8B-SFT}} \\
        \cmidrule(lr){2-4}\cmidrule(lr){5-6}\cmidrule(lr){7-8}
        & \textbf{Token} & \textbf{Context} & \textbf{Cross-fitting}
        & \textbf{AIME24} & \textbf{HMMT25}
        & \textbf{AIME24} & \textbf{HMMT25} \\
        \midrule
        Lightning OPD
        & \xmark & \xmark & \xmark
        & 59.6 & 34.6
        & 80.6 & 62.9 \\
        Context only
        & \xmark & \cmark & \cmark
        & 61.5 & 36.6
        & 82.3 & 62.6 \\
        Token only
        & \cmark & \xmark & \cmark
        & 61.3 & 35.4
        & 82.1 & 61.8 \\
        In-sample estimation
        & \cmark & \cmark & \xmark
        & 62.7 & \textbf{36.7}
        & 81.8 & 63.1 \\
        \addlinespace[1pt]
        \rowcolor{mitblue}
        \framework
        & \cmark & \cmark & \cmark
        & \textbf{63.5} & \textbf{36.7}
        & \textbf{82.4} & \textbf{63.8} \\
        \bottomrule
    \end{tabular}
\end{table*}

\section{Conclusion}
\label{sec:conclusion}

Teacher consistency is central to effective OPD, yet it is often unavailable when an SFT reference and a later distillation teacher are selected independently. We study this cross-teacher setting and find that chosen-token disagreement contains potentially useful context-specific teacher evidence together with a recurring component that is predictable across cached rollouts and associated with differences in wording, formatting, and reasoning cadence. \framework estimates this component from other cached rollouts as an operational proxy for style-token bias and subtracts it before the reference-anchored offline OPD update, leaving the residual signal to drive distillation. Across the Qwen3-4B-SFT and Klear-Reasoner-8B-SFT settings, \framework delivers consistent improvements over Lightning OPD on both mathematical reasoning and code generation tasks. While our evaluation is limited to these tasks and Qwen-family models, the consistent gains establish Lightning OPD 2.0 as a practical framework for efficient and effective cross-teacher OPD, enabling the SFT reference and distillation teacher to be selected independently.

{
    \small
    \bibliographystyle{plainnat}
    \bibliography{arxiv}

@inproceedings{ouyang2022training,
  title={Training language models to follow instructions with human feedback},
  author={Ouyang, Long and Wu, Jeff and Jiang, Xu and Almeida, Diogo and Wainwright, Carroll and Mishkin, Pamela and Zhang, Chong and Agarwal, Sandhini and Slama, Katarina and Ray, Alex and others},
  booktitle={Advances in Neural Information Processing Systems},
  volume={35},
  year={2022}
}

@article{hu2025openreasoner,
  title={Open-Reasoner-Zero: An Open Source Approach to Scaling Up Reinforcement Learning on the Base Model},
  author={Hu, Jingcheng and Zhang, Yinmin and Han, Qi and Jiang, Daxin and Zhang, Xiangyu and Shum, Heung-Yeung},
  journal={arXiv preprint arXiv:2503.24290},
  year={2025}
}

@article{guo2025deepseek,
  title={Deepseek-r1: Incentivizing reasoning capability in llms via reinforcement learning},
  author={Guo, Daya and Yang, Dejian and Zhang, Haowei and Song, Junxiao and Wang, Peiyi and Zhu, Qihao and Xu, Runxin and Zhang, Ruoyu and Ma, Shirong and Bi, Xiao and others},
  journal={arXiv preprint arXiv:2501.12948},
  year={2025}
}

@misc{deepseek2025r10528,
  title={{DeepSeek-R1-0528}},
  author={{DeepSeek-AI}},
  howpublished={\url{https://huggingface.co/deepseek-ai/DeepSeek-R1-0528}},
  year={2025}
}

@article{team2026kimi,
  title={Kimi K2. 5: Visual Agentic Intelligence},
  author={Team, Kimi and Bai, Tongtong and Bai, Yifan and Bao, Yiping and Cai, SH and Cao, Yuan and Charles, Y and Che, HS and Chen, Cheng and Chen, Guanduo and others},
  journal={arXiv preprint arXiv:2602.02276},
  year={2026}
}

@article{singh2025openai,
  title={Openai gpt-5 system card},
  author={Singh, Aaditya and Fry, Adam and Perelman, Adam and Tart, Adam and Ganesh, Adi and El-Kishky, Ahmed and McLaughlin, Aidan and Low, Aiden and Ostrow, AJ and Ananthram, Akhila and others},
  journal={arXiv preprint arXiv:2601.03267},
  year={2025}
}

@article{nvidia_nemotron_3_2025,
  title={NVIDIA Nemotron 3: Efficient and Open Intelligence},
  author={{NVIDIA}},
  journal={arXiv preprint arXiv:2512.20856},
  year={2025}
}

@article{xiao2026mimo,
  title={Mimo-v2-flash technical report},
  author={Xiao, Bangjun and Xia, Bingquan and Yang, Bo and Gao, Bofei and Shen, Bowen and Zhang, Chen and He, Chenhong and Lou, Chiheng and Luo, Fuli and Wang, Gang and others},
  journal={arXiv preprint arXiv:2601.02780},
  year={2026}
}

@article{team2025kimik15,
  title={Kimi k1.5: Scaling reinforcement learning with {LLMs}},
  author={Team, Kimi and Du, Angang and Gao, Bofei and Xing, Bowei and Jiang, Changjiu and others},
  journal={arXiv preprint arXiv:2501.12599},
  year={2025}
}

@article{liu2024deepseekv3,
  title={{DeepSeek-V3} technical report},
  author={Liu, Aixin and Feng, Bei and Xue, Bing and Wang, Bingxuan and Wu, Bochao and Lu, Chengda and Zhao, Chengqi and Deng, Chenggang and Zhang, Chengpeng and others},
  journal={arXiv preprint arXiv:2412.19437},
  year={2024}
}

@article{yang2024qwen25math,
  title={{Qwen2.5-Math} technical report: Toward mathematical expert model via self-improvement},
  author={Yang, An and Zhang, Beichen and Hui, Binyuan and Gao, Bofei and Yu, Bowen and Li, Chengpeng and Liu, Dayiheng and Tu, Jianhong and Zhou, Jingren and Lin, Junyang and others},
  journal={arXiv preprint arXiv:2409.12122},
  year={2024}
}

@article{guha2025openthoughts,
  title={Openthoughts: Data recipes for reasoning models},
  author={Guha, Etash and Marten, Ryan and Keh, Sedrick and Raoof, Negin and Smyrnis, Georgios and Bansal, Hritik and Nezhurina, Marianna and Mercat, Jean and Vu, Trung and Sprague, Zayne and others},
  journal={arXiv preprint arXiv:2506.04178},
  year={2025}
}

@article{schulman2017proximal,
  title={Proximal policy optimization algorithms},
  author={Schulman, John and Wolski, Filip and Dhariwal, Prafulla and Radford, Alec and Klimov, Oleg},
  journal={arXiv preprint arXiv:1707.06347},
  year={2017}
}

@article{shao2024deepseekmath,
  title={Deepseekmath: Pushing the limits of mathematical reasoning in open language models},
  author={Shao, Zhihong and Wang, Peiyi and Zhu, Qihao and Xu, Runxin and Song, Junxiao and Bi, Xiao and Zhang, Haowei and Zhang, Mingchuan and Li, YK and Wu, Yang and others},
  journal={arXiv preprint arXiv:2402.03300},
  year={2024}
}

@article{hu2025reinforcepp,
  title={{REINFORCE++}: A Simple and Efficient Approach for Aligning Large Language Models},
  author={Hu, Jian},
  journal={arXiv preprint arXiv:2501.03262},
  year={2025}
}

@article{ahmadian2024back,
  title={Back to Basics: Revisiting {REINFORCE}-Style Optimization for Learning from Human Feedback in {LLMs}},
  author={Ahmadian, Arash and Cremer, Chris and Gall{\'e}, Matthias and Fadaee, Marzieh and Kreutzer, Julia and Pietquin, Olivier and {\"U}st{\"u}n, Ahmet and Hooker, Sara},
  journal={arXiv preprint arXiv:2402.14740},
  year={2024}
}

@article{li2023remax,
  title={{ReMax}: A Simple, Effective, and Efficient Reinforcement Learning Method for Aligning Large Language Models},
  author={Li, Ziniu and Xu, Tian and Zhang, Yushun and Yu, Yang and Sun, Ruoyu and Luo, Zhi-Quan},
  journal={arXiv preprint arXiv:2310.10505},
  year={2023}
}

@article{liu2025drgrpo,
  title={Understanding {R1-Zero}-Like Training: A Critical Perspective},
  author={Liu, Zichen and Chen, Changyu and Li, Wenjun and Qi, Penghui and Pang, Tianyu and Du, Chao and Lee, Wee Sun and Lin, Min},
  journal={arXiv preprint arXiv:2503.20783},
  year={2025}
}

@article{zheng2025gspo,
  title={Group Sequence Policy Optimization},
  author={Zheng, Chujie and Liu, Shixuan and Li, Mingze and Chen, Xiong-Hui and Yu, Bowen and Gao, Chang and Dang, Kai and Liu, Yuqiong and Men, Rui and Yang, An and Zhou, Jingren and Lin, Junyang},
  journal={arXiv preprint arXiv:2507.18071},
  year={2025}
}

@article{minimax2025cispo,
  title={{MiniMax-M1}: Scaling Test-Time Compute Efficiently with Lightning Attention},
  author={MiniMax and Shan, Aonian and Gong, Bangwei and Yang, Bo and others},
  journal={arXiv preprint arXiv:2506.13585},
  year={2025}
}

@article{yuan2025vcppo,
  title={What's Behind {PPO}'s Collapse in Long-{CoT}? Value Optimization Holds the Secret},
  author={Yuan, Yufeng and Yue, Yu and Zhu, Ruofei and Fan, Tiantian and Yan, Lin},
  journal={arXiv preprint arXiv:2503.01491},
  year={2025}
}

@article{yue2025vapo,
  title={{VAPO}: Efficient and Reliable Reinforcement Learning for Advanced Reasoning Tasks},
  author={Yue, Yu and Yuan, Yufeng and Yu, Qiying and Zuo, Xiaochen and Zhu, Ruofei and Xu, Wenyuan and Chen, Jiaze and Wang, Chengyi and Fan, Tiantian and Du, Zhengyin and Wei, Xiangpeng and others},
  journal={arXiv preprint arXiv:2504.05118},
  year={2025}
}

@article{kazemnejad2024vineppo,
  title={{VinePPO}: Refining Credit Assignment in {RL} Training of {LLMs}},
  author={Kazemnejad, Amirhossein and Aghajohari, Milad and Portelance, Eva and Sordoni, Alessandro and Reddy, Siva and Courville, Aaron and Le Roux, Nicolas},
  journal={arXiv preprint arXiv:2410.01679},
  year={2024}
}

@article{yu2025dapo,
  title={Dapo: An open-source llm reinforcement learning system at scale},
  author={Yu, Qiying and Zhang, Zheng and Zhu, Ruofei and Yuan, Yufeng and Zuo, Xiaochen and Yue, Yu and Dai, Weinan and Fan, Tiantian and Liu, Gaohong and Liu, Lingjun and others},
  journal={arXiv preprint arXiv:2503.14476},
  year={2025}
}

@article{ye2025blackbox,
  title={Black-box on-policy distillation of large language models},
  author={Ye, Tianzhu and Dong, Li and Chi, Zewen and Wu, Xun and Huang, Shaohan and Wei, Furu},
  journal={arXiv preprint arXiv:2511.10643},
  year={2025}
}

@article{cui2025prime,
  title={Process reinforcement through implicit rewards},
  author={Cui, Ganqu and Yuan, Lifan and Wang, Zefan and Wang, Hanbin and Zhang, Yuchen and Chen, Jiacheng and Li, Wendi and He, Bingxiang and Fan, Yuchen and Yu, Tianyu and others},
  journal={arXiv preprint arXiv:2502.01456},
  year={2025}
}

@inproceedings{kim2016sequence,
  title={Sequence-level knowledge distillation},
  author={Kim, Yoon and Rush, Alexander M},
  booktitle={Proceedings of the 2016 Conference on Empirical Methods in Natural Language Processing},
  year={2016}
}

@article{hinton2015distilling,
  title={Distilling the knowledge in a neural network},
  author={Hinton, Geoffrey and Vinyals, Oriol and Dean, Jeff},
  journal={NIPS Deep Learning Workshop},
  year={2015}
}

@inproceedings{gu2024minillm,
  title={{MiniLLM}: Knowledge distillation of large language models},
  author={Gu, Yuxian and Dong, Li and Wei, Furu and Huang, Minlie},
  booktitle={The Twelfth International Conference on Learning Representations},
  year={2024}
}

@article{lu2025onpolicydistillation,
  author={Kevin Lu and Thinking Machines Lab},
  title={On-Policy Distillation},
  journal={Thinking Machines Lab: Connectionism},
  year={2025},
  note={https://thinkingmachines.ai/blog/on-policy-distillation},
  doi={10.64434/tml.20251026},
}

@inproceedings{agarwal2024policy,
  title={On-policy distillation of language models: Learning from self-generated mistakes},
  author={Agarwal, Rishabh and Vieillard, Nino and Zhou, Yongchao and Stanczyk, Piotr and Garea, Sabela Ramos and Geist, Matthieu and Bachem, Olivier},
  booktitle={The Twelfth International Conference on Learning Representations},
  year={2024}
}

@article{yang2026learning,
  title={Learning beyond Teacher: Generalized On-Policy Distillation with Reward Extrapolation},
  author={Yang, Wenkai and Liu, Weijie and Xie, Ruobing and Yang, Kai and Yang, Saiyong and Lin, Yankai},
  journal={arXiv preprint arXiv:2602.12125},
  year={2026}
}

@article{yang2025qwen3,
  title={Qwen3 technical report},
  author={Yang, An and Li, Anfeng and Yang, Baosong and Zhang, Beichen and Hui, Binyuan and Zheng, Bo and Yu, Bowen and Gao, Chang and Huang, Chengen and Lv, Chenxu and others},
  journal={arXiv preprint arXiv:2505.09388},
  year={2025}
}

@article{zhao2026self,
  title={Self-Distilled Reasoner: On-Policy Self-Distillation for Large Language Models},
  author={Zhao, Siyan and Xie, Zhihui and Liu, Mengchen and Huang, Jing and Pang, Guan and Chen, Feiyu and Grover, Aditya},
  journal={arXiv preprint arXiv:2601.18734},
  year={2026}
}

@article{shenfeld2025razor,
  title={RL's Razor: Why Online Reinforcement Learning Forgets Less},
  author={Shenfeld, Idan and Pari, Jyothish and Agrawal, Pulkit},
  journal={arXiv preprint arXiv:2509.04259},
  year={2025}
}

@article{shenfeld2026self,
  title={Self-Distillation Enables Continual Learning},
  author={Shenfeld, Idan and Damani, Mehul and H{\"u}botter, Jonas and Agrawal, Pulkit},
  journal={arXiv preprint arXiv:2601.19897},
  year={2026}
}

@article{hubotter2026reinforcement,
  title={Reinforcement Learning via Self-Distillation},
  author={H{\"u}botter, Jonas and L{\"u}beck, Frederike and Behric, Lejs and Baumann, Anton and Bagatella, Marco and Marta, Daniel and Hakimi, Ido and Shenfeld, Idan and Buening, Thomas Kleine and Guestrin, Carlos and others},
  journal={arXiv preprint arXiv:2601.20802},
  year={2026}
}

@article{Nemotron_Cascade,
  title={Nemotron-Cascade: Scaling Cascaded Reinforcement Learning for General-Purpose Reasoning Models},
  author={Chen, Yang and Yang, Zhuolin and Liu, Zihan and Dai, Wenliang and Wang, Boxin and Lin, Sheng-Chieh and Lee, Chankyu and Shoeybi, Mohammad and Catanzaro, Bryan and Ping, Wei},
  journal={arXiv preprint arXiv:2512.13607},
  year={2025}
}

@article{Nemotron_Cascade_2,
  title={Nemotron-Cascade 2: Post-Training LLMs with Cascade RL and Multi-Domain On-Policy Distillation},
  author={Yang, Zhuolin and Liu, Zihan and Chen, Yang and Dai, Wenliang and Wang, Boxin and Lin, Sheng-Chieh and Lee, Chankyu and Chen, Yangyi and Jiang, Dongfu and He, Jiafan and Pi, Renjie and Lam, Grace and Lee, Nayeon and Bukharin, Alexander and Shoeybi, Mohammad and Catanzaro, Bryan and Ping, Wei},
  journal={arXiv preprint arXiv:2603.19220},
  year={2026}
}

@article{penaloza2026privileged,
  title={Privileged Information Distillation for Language Models},
  author={Penaloza, Emiliano and Vattikonda, Dheeraj and Gontier, Nicolas and Lacoste, Alexandre and Charlin, Laurent and Caccia, Massimo},
  journal={arXiv preprint arXiv:2602.04942},
  year={2026}
}

@article{liang2026orbit,
  title={ORBIT: On-Policy Exploration-Exploitation for Controllable Multi-Budget Reasoning},
  author={Liang, Kun and Bai, Clive and Xu, Xin and Tang, Chenming and Lee, Sanwoo and Liu, Weijie and Yang, Saiyong and Wu, Yunfang},
  journal={arXiv preprint arXiv:2601.08310},
  year={2026}
}

@inproceedings{xu2025speculative,
  title={Speculative Knowledge Distillation: Bridging the Teacher-Student Gap Through Interleaved Sampling},
  author={Xu, Wenda and Han, Rujun and Wang, Zifeng and Le, Long and Madeka, Dhruv and Li, Lei and Wang, William Yang and Agarwal, Rishabh and Lee, Chen-Yu and Pfister, Tomas},
  booktitle={The Thirteenth International Conference on Learning Representations},
  year={2025}
}

@article{ke2026respecting,
  title={Respecting Self-Uncertainty in On-Policy Self-Distillation for Efficient {LLM} Reasoning},
  author={Ke, Junlong and Wen, Zichen and Li, Weijia and He, Conghui and Zhang, Linfeng},
  journal={arXiv preprint arXiv:2605.13255},
  year={2026}
}

@article{armandpour2026unmasking,
  title={Unmasking On-Policy Distillation: Where It Helps, Where It Hurts, and Why},
  author={Armandpour, Mohammadreza and Ilhan, Fatih and Harrison, David and Jaiswal, Ajay and Hoang, Duc N. M. and Faghri, Fartash and Zhang, Yizhe and Cho, Minsik and Farajtabar, Mehrdad},
  journal={arXiv preprint arXiv:2605.10889},
  year={2026}
}

@article{xu2026sign,
  title={{SG-OPD}: Sign-Gated On-Policy Distillation via Sign-Consistency Gating and Phased Teacher Sampling},
  author={Xu, Haoran and Wang, Hongyu and Gao, Yifei and Li, Jiaze and Zhang, Xiaofeng and Yuan, Xiaosong},
  journal={arXiv preprint arXiv:2606.09304},
  year={2026}
}

@article{tan2026selfsupervised,
  title={Self-Supervised On-Policy Distillation for Reasoning Language Models},
  author={Tan, Zhiquan and Hong, Yinrong},
  journal={arXiv preprint arXiv:2605.17497},
  year={2026}
}

@article{yang2026ogls,
  title={{OGLS-SD}: On-Policy Self-Distillation with Outcome-Guided Logit Steering for {LLM} Reasoning},
  author={Yang, Yuxiao and Wang, Xiaoyun and Zhang, Weitong},
  journal={arXiv preprint arXiv:2605.12400},
  year={2026}
}

@article{dong2023raft,
  title={{RAFT}: Reward ranked finetuning for generative foundation model alignment},
  author={Dong, Hanze and Xiong, Wei and Goyal, Deepanshu and Zhang, Yihan and Chow, Winnie and Pan, Rui and Diao, Shizhe and Zhang, Jipeng and Shum, Kashun and Zhang, Tong},
  journal={Transactions on Machine Learning Research},
  year={2023}
}

@inproceedings{rafailov2023direct,
  title={Direct preference optimization: Your language model is secretly a reward model},
  author={Rafailov, Rafael and Sharma, Archit and Mitchell, Eric and Manning, Christopher D and Ermon, Stefano and Finn, Chelsea},
  booktitle={Advances in Neural Information Processing Systems},
  volume={36},
  year={2023}
}

@article{chen2025bridge,
  title={Beyond Two-Stage Training: Cooperative {SFT} and {RL} for {LLM} Reasoning},
  author={Chen, Liang and Han, Xueting and Shen, Li and Bai, Jing and Wong, Kam-Fai},
  journal={arXiv preprint arXiv:2509.06948},
  year={2025}
}

@article{liu2025uft,
  title={{UFT}: Unifying Supervised and Reinforcement Fine-Tuning},
  author={Liu, Mingyang and Farina, Gabriele and Ozdaglar, Asuman},
  journal={arXiv preprint arXiv:2505.16984},
  year={2025}
}

@article{huang2025blend,
  title={Blending Supervised and Reinforcement Fine-Tuning with Prefix Sampling},
  author={Huang, Zeyu and Cheng, Tianhao and Qiu, Zihan and Wang, Zili and Xu, Yinghui and Ponti, Edoardo M. and Titov, Ivan},
  journal={arXiv preprint arXiv:2507.01679},
  year={2025}
}

@misc{slime2025,
  title={slime: An {SGLang}-Native Post-Training Framework for {RL} Scaling},
  author={{THUDM}},
  year={2025},
  howpublished={\url{https://github.com/THUDM/slime}}
}

@misc{aimo2024aime,
  title={{AIME} 2024},
  author={{AI-MO}},
  year={2024},
  howpublished={\url{https://huggingface.co/datasets/AI-MO/aimo-validation-aime}}
}

@misc{opencompass2025aime,
  title={{AIME} 2025},
  author={{OpenCompass}},
  year={2025},
  howpublished={\url{https://github.com/open-compass/opencompass}}
}

@article{balunovic2025matharena,
  title={{MathArena}: Evaluating {LLMs} on Uncontaminated Math Competitions},
  author={Balunovi{\'c}, Mislav and Dekoninck, Jasper and Petrov, Ivo and Jovanovi{\'c}, Nikola and Vechev, Martin},
  journal={arXiv preprint arXiv:2505.23281},
  year={2025}
}

@article{jain2024livecodebench,
  title={{LiveCodeBench}: Holistic and Contamination Free Evaluation of Large Language Models for Code},
  author={Jain, Naman and Han, King and Gu, Alex and Li, Wen-Ding and Yan, Fanjia and Zhang, Tianjun and Wang, Sida and Solar-Lezama, Armando and Sen, Koushik and Stoica, Ion},
  journal={arXiv preprint arXiv:2403.07974},
  year={2024}
}

@article{yan2025luffy,
  title={Learning to Reason under Off-Policy Guidance},
  author={Yan, Jianhao and Li, Yafu and Hu, Zican and Wang, Zhi and Cui, Ganqu and Qu, Xiaoye and Cheng, Yu and Zhang, Yue},
  journal={arXiv preprint arXiv:2504.14945},
  year={2025}
}

@article{ma2026relift,
  title={Learning What Reinforcement Learning Can't: Interleaved Online Fine-Tuning for Hardest Questions},
  author={Ma, Lu and Liang, Hao and Qiang, Meiyi and Tang, Lexiang and Ma, Xiaochen and Wong, Zhen Hao and Niu, Junbo and Shen, Chengyu and He, Runming and Li, Yanhao and Cui, Bin and Zhang, Wentao},
  journal={arXiv preprint arXiv:2506.07527},
  year={2026}
}

@article{zhang2025chord,
  title={On-Policy {RL} Meets Off-Policy Experts: Harmonizing Supervised Fine-Tuning and Reinforcement Learning via Dynamic Weighting},
  author={Zhang, Wenhao and Xie, Yuexiang and Sun, Yuchang and Chen, Yanxi and Wang, Guoyin and Li, Yaliang and Ding, Bolin and Zhou, Jingren},
  journal={arXiv preprint arXiv:2508.11408},
  year={2025}
}

@article{yue2025does,
  title={Does reinforcement learning really incentivize reasoning capacity in llms beyond the base model?},
  author={Yue, Yang and Chen, Zhiqi and Lu, Rui and Zhao, Andrew and Wang, Zhaokai and Song, Shiji and Huang, Gao},
  journal={arXiv preprint arXiv:2504.13837},
  year={2025}
}

@article{song2026survey,
  title={A Survey of On-Policy Distillation for Large Language Models},
  author={Song, Mingyang and Zheng, Mao},
  journal={arXiv preprint arXiv:2604.00626},
  year={2026}
}

@article{chen2026rethinking,
  title={Rethinking On-Policy Distillation of Large Language Models: Phenomenology, Mechanism, and Recipe},
  author={Li, Yaxuan and Zuo, Yuxin and He, Bingxiang and Zhang, Jinqian and Xiao, Chaojun and Qian, Cheng and Yu, Tianyu and Gao, Huan-ang and Yang, Wenkai and Liu, Zhiyuan and Ding, Ning},
  journal={arXiv preprint arXiv:2604.13016},
  year={2026}
}

@article{fu2026revisiting,
  title={Revisiting On-Policy Distillation: Empirical Failure Modes and Simple Fixes},
  author={Fu, Yuqian and Huang, Haohuan and Jiang, Kaiwen and Zhu, Yuanheng and Zhao, Dongbin},
  journal={arXiv preprint arXiv:2603.25562},
  year={2026}
}

@article{wang2026entropy,
  title={Entropy-Aware On-Policy Distillation of Language Models},
  author={Jin, Woogyeol and Min, Taywon and Yang, Yongjin and Kadhe, Swanand Ravindra and Zhou, Yi and Wei, Dennis and Baracaldo, Nathalie and Lee, Kimin},
  journal={arXiv preprint arXiv:2603.07079},
  year={2026}
}

@article{xu2026reinforcement,
  title={Reinforcement-Aware Knowledge Distillation for {LLM} Reasoning},
  author={Zhang, Zhaoyang and Jiang, Shuli and Shen, Yantao and Zhang, Yuting and Ram, Dhananjay and Yang, Shuo and Tu, Zhuowen and Xia, Wei and Soatto, Stefano},
  journal={arXiv preprint arXiv:2602.22495},
  year={2026}
}

@article{zeng2026glm5,
  title={{GLM-5}: From Vibe Coding to Agentic Engineering},
  author={Zeng, Aohan and others},
  journal={arXiv preprint},
  year={2026}
}

@article{rang2025revealing,
  title={Revealing the Power of Post-Training for Small Language Models via Knowledge Distillation},
  author={Rang, Miao and Bi, Zhenni and Zhou, Hang and Chen, Hanting and Xiao, An and Guo, Tianyu and Han, Kai and Chen, Xinghao and Wang, Yunhe},
  journal={arXiv preprint arXiv:2509.26497},
  year={2025}
}

@article{wu2026lightning,
  title={Lightning {OPD}: Efficient Post-Training for Large Reasoning Models with Offline On-Policy Distillation},
  author={Wu, Yecheng and Han, Song and Cai, Han},
  journal={arXiv preprint arXiv:2604.13010},
  year={2026}
}

@article{xie2026position,
  title={On the Position Bias of On-Policy Distillation},
  author={Xie, Yan and Zhu, Sijie and Wen, Tiansheng and Chen, Bo and Wang, Yifei},
  journal={arXiv preprint arXiv:2606.22600},
  year={2026}
}

@article{wang2026teachability,
  title={Not All Disagreement Is Learnable: Token Teachability in On-Policy Distillation},
  author={Wang, Yuanyi and Lu, Su and Gu, Yanggan and Wang, Pengkai and Yang, Yifan and Yan, Zhaoyi and Xie, Congkai and Wu, Jianmin and Yang, Hongxia},
  journal={arXiv preprint arXiv:2605.26844},
  year={2026}
}

@misc{kwai2026klear,
  title={{Klear-Reasoner-8B-SFT}},
  author={{Kwai-Klear}},
  howpublished={\url{https://huggingface.co/Kwai-Klear/Klear-Reasoner-8B-SFT}},
  year={2026}
}

@article{wilcox2023testing,
  title={Testing the Predictions of Surprisal Theory in 11 Languages},
  author={Wilcox, Ethan G. and Pimentel, Tiago and Meister, Clara and Cotterell, Ryan and Levy, Roger P.},
  journal={Transactions of the Association for Computational Linguistics},
  volume={11},
  pages={1451--1470},
  year={2023},
  doi={10.1162/tacl_a_00612}
}

@article{chernozhukov2018double,
  title={Double/Debiased Machine Learning for Treatment and Structural Parameters},
  author={Chernozhukov, Victor and Chetverikov, Denis and Demirer, Mert and Duflo, Esther and Hansen, Christian and Newey, Whitney and Robins, James},
  journal={The Econometrics Journal},
  volume={21},
  number={1},
  pages={C1--C68},
  year={2018},
  doi={10.1111/ectj.12097}
}

@inproceedings{guo2021machine,
  title={Machine Learning for Variance Reduction in Online Experiments},
  author={Guo, Yongyi and Coey, Dominic and Konutgan, Mikael and Li, Wenting and Schoener, Chris and Goldman, Matt},
  booktitle={Advances in Neural Information Processing Systems},
  volume={34},
  year={2021}
}
}

\clearpage
\appendix

\end{document}